\newif\ifonecolumn

\ifonecolumn 
    \documentclass[12pt,journal,a4paper,oneside,onecolumn]{IEEEtran}
\else 
    \documentclass[10pt,journal,final,a4paper,oneside,twocolumn]{IEEEtran}
\fi

\usepackage{cite} 
\usepackage{graphicx} 
\usepackage{hyperref}
\hypersetup{colorlinks=true, allcolors=blue}
\usepackage{comment}
\usepackage{scalefnt}

\usepackage{amssymb}
\usepackage{amsmath}
\usepackage{dsfont}
\usepackage{slashbox}
\usepackage{url}
\usepackage{caption}
\usepackage{subcaption}
\usepackage{multirow}
\usepackage{booktabs}
\usepackage{hyphenat}
\usepackage{diagbox}
\usepackage{makecell}
\usepackage{tabularray}

\usepackage[linesnumbered,ruled,vlined]{algorithm2e}

\usepackage{fancyhdr}
\renewcommand{\today}{\number\year /\number\month /\number\day}
\makeatletter
\let\oldlt\longtable
\let\endoldlt\endlongtable
\def\longtable{\@ifnextchar[\longtable@i \longtable@ii}
\def\longtable@i[#1]{\begin{figure}[t]
\onecolumn
\begin{minipage}{0.5\textwidth}
\oldlt[#1]
}
\def\longtable@ii{\begin{figure}[t]
\onecolumn
\begin{minipage}{0.5\textwidth}
\oldlt
}
\def\endlongtable{\endoldlt
\end{minipage}
\twocolumn
\end{figure}}
\makeatother

\makeatletter

\newcommand{\Rmnum}[1]{\expandafter\@slowromancap\romannumeral #1@}
\let\oldnl\nl
\newcommand{\nonl}{\renewcommand{\nl}{\let\nl\oldnl}}

\input{epsf}

\IEEEoverridecommandlockouts

\begin{document}

\definecolor{mygray}{gray}{.9}
\definecolor{mypink}{rgb}{.99,.91,.95}
\definecolor{myblue}{rgb}{.35,.67,.99}
\def\dis{\strut\displaystyle}

\title{Lightweight Spatio-Temporal Attention Network with Graph Embedding and Rotational Position Encoding for Traffic Forecasting}

\author{Xiao Wang and Shun-Ren~Yang,~\IEEEmembership{Member,~IEEE}
\thanks{X. Wang and S.-R. Yang are with the Department of Computer Science and the Institute of Communications Engineering, National Tsing Hua University, Hsinchu 30013, Taiwan (e-mail: sryang@cs.nthu.edu.tw).}
}

\maketitle

\begin{abstract}
Traffic forecasting is a key task in the field of Intelligent Transportation Systems. Recent research on traffic forecasting has mainly focused on combining graph neural networks (GNNs) with other models. However, GNNs only consider short-range spatial information. In this study, we present a novel model termed ‌LSTAN-GERPE‌ (Lightweight Spatio-Temporal Attention Network with Graph Embedding and Rotational Position Encoding). This model leverages both Temporal and Spatial Attention mechanisms to effectively capture long-range traffic dynamics. Additionally, the optimal frequency for rotational position encoding is determined through a grid search approach in both the spatial and temporal attention mechanisms. This systematic optimization enables the model to effectively capture complex traffic patterns. The model also enhances feature representation by incorporating geographical location maps into the spatio-temporal embeddings. Without extensive feature engineering, the proposed method in this paper achieves advanced accuracy on the real-world traffic forecasting datasets PeMS04 and PeMS08.
\end{abstract}

\begin{IEEEkeywords}
Traffic Forecast, Spatio-Temporal Attention, RoPE, Eigenvector
\end{IEEEkeywords}

\section{Introduction}
\begin{figure}[htbp]
    \centering
    \begin{subfigure}[b]{0.23\textwidth} 
        \includegraphics[width=\textwidth]{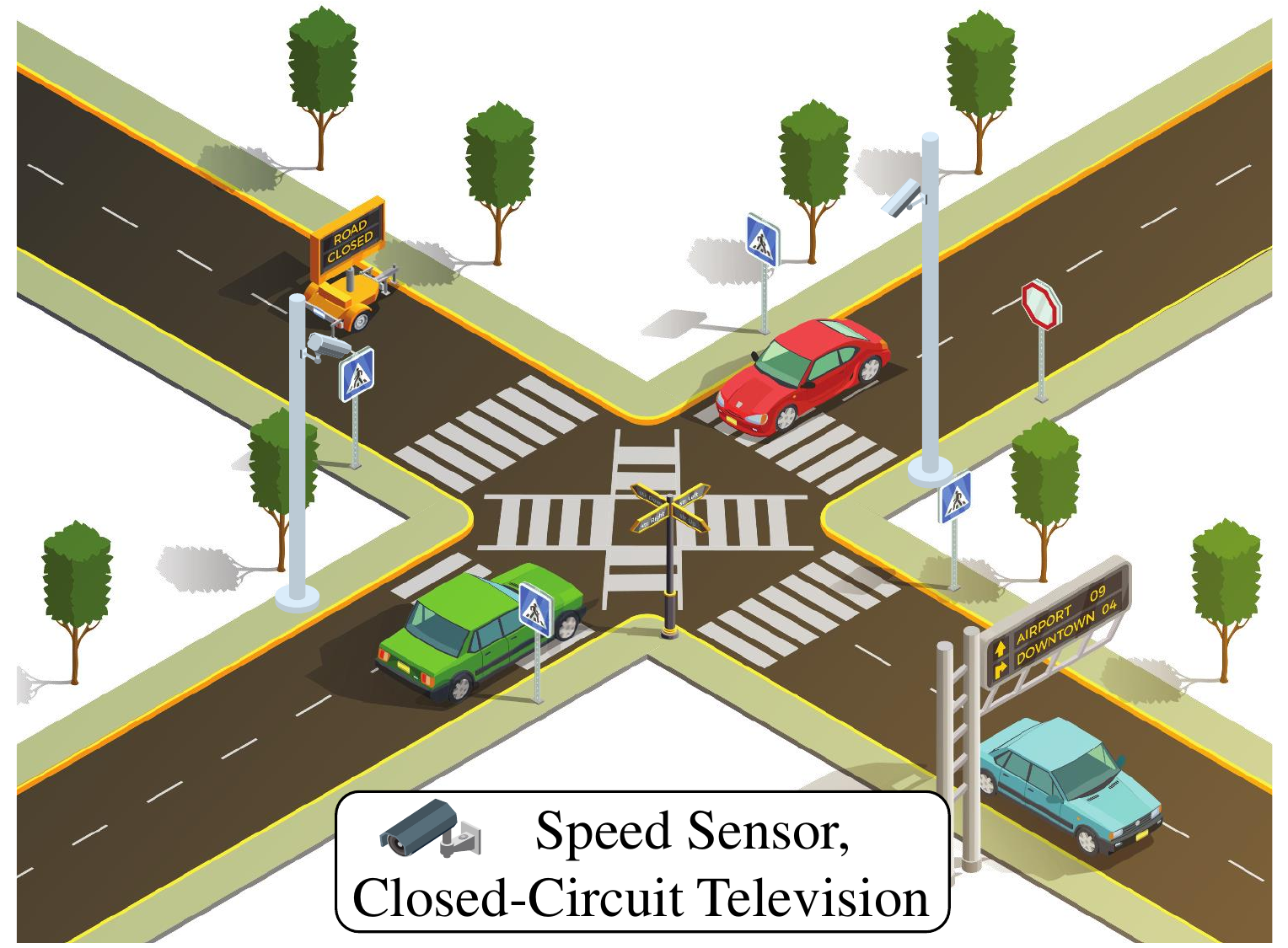}
        \caption{Traffic Monitoring Systems at Intersections}
        \label{fig:traffic_monitoring}
    \end{subfigure}
    \hfill 
    \begin{subfigure}[b]{0.23\textwidth} 
        \includegraphics[width=\textwidth]{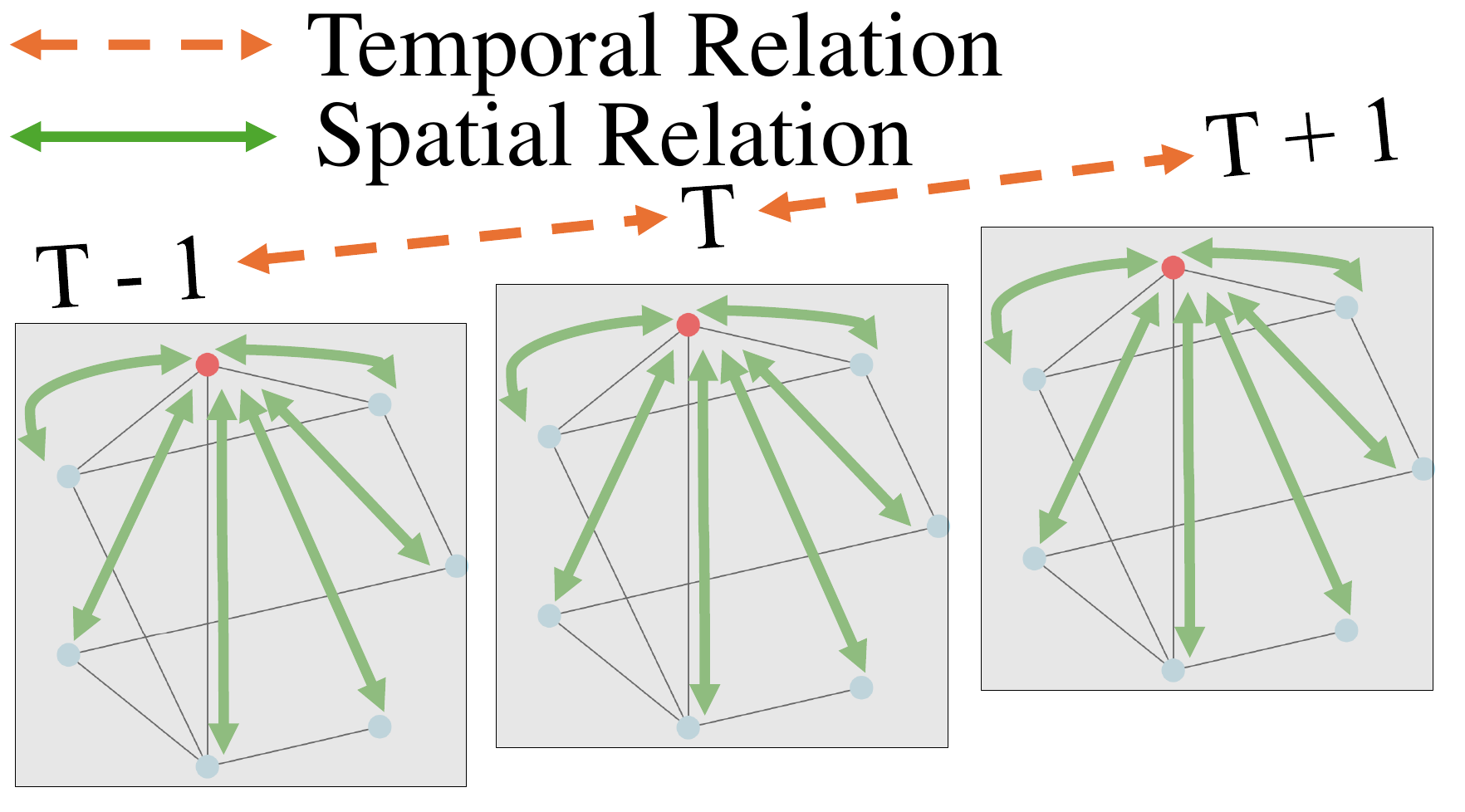}
        \caption{Traffic Data Spatio-Temporal Dynamics}
        \label{fig:traffic_st}
    \end{subfigure}
    \vskip\baselineskip 

    \begin{subfigure}[b]{0.48\textwidth} 
        \includegraphics[width=\textwidth]{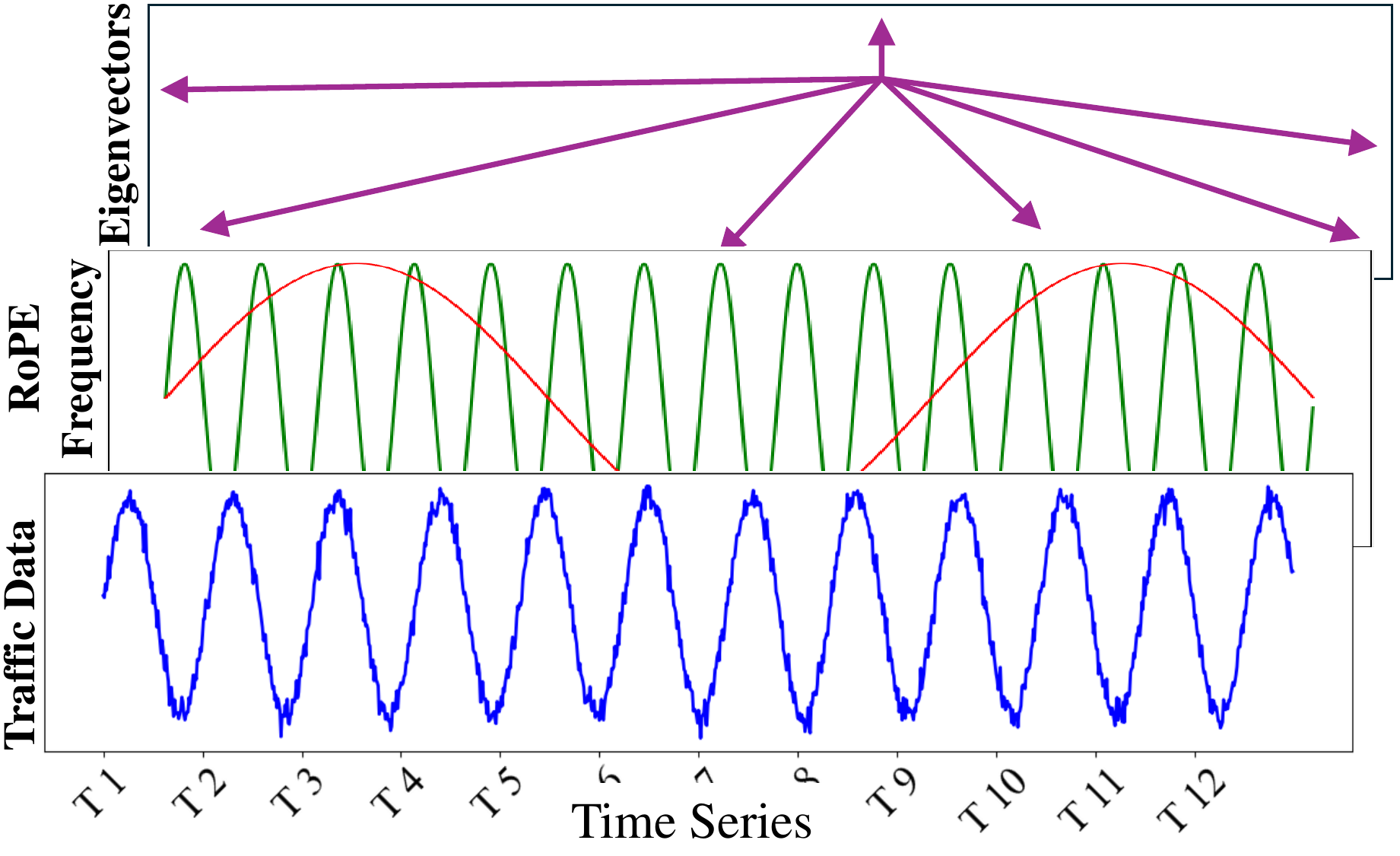}
        \caption{Composite Traffic Pattern Representation}
        \label{fig:traffic_pattern}
    \end{subfigure}
    \caption{Traffic Data Analysis and Monitoring Suite}
    \label{fig:traffic_all}
\end{figure}

With the rapid development of Intelligent Transportation Systems, traffic forecasting plays a crucial role in urban planning and traffic management. As depicted in Fig. \ref{fig:traffic_monitoring}, the process begins with traffic data collection at road intersections by monitors. Subsequently, these data are harnessed for forecasting within smart traffic systems, with the analytical results potentially directing vehicle rerouting to bolster traffic efficiency. In the realm of traffic forecasting, deep learning-based methods, including Recurrent Neural Networks (RNNs), Convolutional Neural Networks (CNNs), and Graph Neural Networks (GNNs), have been extensively utilized to model the temporal and spatial correlations inherent in traffic forecasting tasks.

GNNs have recently gained significant attention due to their ability to handle graph data. Road networks naturally exist in the form of graphs. Graph Convolutional Networks (GCNs) represent a specific class of GNNs that capture spatial dependencies between nodes by applying convolutional operations directly on the graph structure. Spatio-temporal GCNs (ST-GCNs)\cite{yu2018spatio} integrate spatial and temporal dimensions, allowing them to capture both spatial dependencies and temporal dynamics of traffic data simultaneously. Spatio-temporal graph attention networks (ST-GANs) introduce an attention mechanism to more flexibly capture dependencies based on the adjacency relationship of nodes.

The Self-Attention mechanism, a popular technique, allows models to automatically focus on important parts of the input sequence. In traffic forecasting, as depicted in Fig. \ref{fig:traffic_st}, this enables models to identify and concentrate on the most critical spatio-temporal relation to predict future traffic conditions. The Traffic Transformer \cite{yan2021learning} employs a transformer encoder to extract and fuse spatial patterns, however, resulting in high model complexity and a large number of parameters.

With the evolution of model architectures, researchers have increasingly recognized the potential of input embedding techniques. For instance, STIM \cite{huang2024high} identifies short-term and long-term patterns, captures periodicity information in the input embedding. However, these techniques can introduce significant computational loads and potential biases.

To address the limitations highlighted above, we have developed the ‌LSTAN-GERPE‌  (standing for Lightweight Spatio-Temporal Attention Network with Graph Embedding and Rotational Position Encoding) model in this paper, which emphasizes continuous optimization of the model architecture without reliance on input embedding-based feature engineering. The main features of our model can be summarized as follows.
\begin{itemize}
 \item Lightweight spatio-temporal attention pair. As correlations exist in Fig. \ref{fig:traffic_st}, we will design a lightweight pair utilizing the spatio-temporal self-attention mechanism which separately extracts the spatial axis' features and temporal axis' features asynchronously for precise prediction of each road's vehicle speeds. We devise a sequential stacking style incorporating spatio-temporal pairs.
\item We leverage Rotary Position Embedding (RoPE) \cite{su2024roformer} to bolster the efficacy of dependency learning and enhance generalization, specifically RoPE across various frequencies to discern patterns in both long and short sequences, as illustrated as the middle block in Fig. \ref{fig:traffic_pattern}. We identify the optimal frequency for spatial and temporal attention mechanisms and subsequently added RoPE to traffic flow data inputs.

\item  Furthermore, we conduct the eigenvalue decomposition on the normalized Laplacian matrix of the road map. As depicted as the bottom layer in Fig. \ref{fig:traffic_pattern}, by appending the eigenvector matrix to the input data, we effectively incorporate the spatial and topological features of the map into our model. 
\end{itemize}

\section{Related Work}
\subsection{Time Series Forecasting Models}
In the early years, traffic forecasting was heavily reliant on traditional mathematical models. For example, historical average models and ARIMA models were used to predict traffic flow. Later, more advanced techniques such as Kalman filters and wavelet neural networks were introduced. In recent years, deep learning-based models have consistently demonstrated superior performance in predicting city-wide traffic flow. Consequently, many scholars now prefer to employ the latest artificial intelligence models for traffic forecasting, as these approaches generally outperform traditional mathematical modeling techniques. For example, \cite{niu2019novel} utilizes a hybrid model that combines convolutional neural networks (CNNs) and long-short-term memory (LSTM) networks.

\subsection{Graph-based method}
\cite{wu2020comprehensive} employs a graph model, STGNN, which integrates GCN and graph attention networks (GAT) to capture spatial relationships and RNN or LSTM to capture temporal dynamics. Unlike GCN, which captures information only from a node's immediate neighbors, GAT captures information from all nodes. ST-GCN \cite{yu2018spatio} further integrates spatial and temporal dimensions, leveraging GCN for spatial dependencies and CNN for temporal dynamics in traffic data. 

\subsection{Self-Attention-based methods}
The Attention Mechanism \cite{vaswani2017attention} is a technique that enables models to automatically focus on important parts of the input sequence. \cite{yan2021learning} employs a transformer encoder to extract and fuse spatial patterns. The HTVGNN \cite{dai2024novel} employs a time-varying mask in its temporal perception multi-head self-attention mechanism to precisely capture temporal dependencies. DTRformer \cite{chen2025dynamic} utilizes the cross-attention mechanism to fuse global temporal and spatial information. STAEformer \cite{liu2023spatio} combine spatio-temporal adaptive embedding and vanilla Transformer. It is applied along the temporal axis as the temporal transformer and along the spatial axis as the spatial transformer.
\cite{jiang2023pdformer} and \cite{huang2024high} utilize self-attention as the model backbone to capture periodicity information in input, identifying short-term and long-term patterns in data embedding. However, these techniques can introduce significant computational load and potential biases due to data feature engineering.
\subsection{Self-Attention Positional Encoding}
\label{sec: Positional Encoding}
Absolute Position Encoding is a method used in Transformer \cite{vaswani2017attention} to incorporate positional information into the input embeddings. For each position \( pos \) in a sequence and each dimension \( i \) of the embedding, the position encoding \( PE(pos, i) \) is computed using sine and cosine functions of different frequencies:
\begin{equation}
\begin{aligned}
PE(pos, 2i) &= \sin\left(\frac{pos}{10000^{2i/d}}\right) \\
PE(pos, 2i+1) &= \cos\left(\frac{pos}{10000^{2i/d}}\right)
\end{aligned}
\end{equation}
where $pos$ is the position in the sequence, $i$ is the dimension index, $d$ is the dimensionality of the embedding space. These position encodings are added to the input embeddings to provide the model with information about the sequence order. However it has limitations in capturing long-range dependencies and may not generalize well to sequences of varying lengths.

To address this limitation, Relative Positional Encoding \cite{shaw2018self} embeds positional information into attention scores by representing positions as relative distances, instead of using absolute positions. Given Query $Q$, Key $K$ and Value $V$ vectors in Transformer, the relative positional encoding R is expressed as:
\begin{align}
\text{Attention}(Q, K) = \text{softmax}\left(\frac{QK^T + R}{\sqrt{d}}\right) V
\end{align}
where $R$ is a relative position matrix representing the relative distances between each position, $d$ is the dimensionality of the embedding space.

RoPE further optimizes relative position encoding by embedding directional information. It is a function $f$ used in Transformer models to dynamically embed positional information into the query and key vectors. The inner product of the query and the key vectors is computed using a function $g$, which solely depends on the relative position $m-n$ beyond the input embeddings.

\begin{align}
\langle f_q(\boldsymbol{x}_m, m), f_k(\boldsymbol{x}_n, n) \rangle = g(\boldsymbol{x}_m, \boldsymbol{x}_n, m-n).
\end{align} 
where $\boldsymbol{x}_m, \boldsymbol{x}_n$ represent embeddings of items in the sequence.

RoPE design the function $f$ as:
\begin{align}
f_{\{q,k\}}(\boldsymbol{x}_m, m) = \boldsymbol{R}_{\Theta,m}^d \boldsymbol{W}_{\{q,k\}} \boldsymbol{x}_m
\end{align}
where
\begin{align}
\small 
\renewcommand{\arraystretch}{0.8} 
\setlength{\arraycolsep}{0.9pt} 
\boldsymbol{R}_{\Theta,m}^d =
\begin{pmatrix}
\cos m\theta_0 & -\sin m\theta_0 & 0 & 0 & \cdots & 0 \\
\sin m\theta_0 & \cos m\theta_0 & 0 & 0 & \cdots & 0 \\
0 & 0 & \cos m\theta_1 & -\sin m\theta_1 & \cdots & 0 \\
0 & 0 & \sin m\theta_1 & \cos m\theta_1 & \cdots & 0 \\
\vdots & \vdots & \vdots & \vdots & \ddots & \vdots \\
0 & 0 & 0 & 0 & \cdots & \cos m\theta_{d/2-1} \\
0 & 0 & 0 & 0 & \cdots & \sin m\theta_{d/2-1}
\end{pmatrix}
\end{align}
where $\Theta$ is the predefined parameters as shown in Equation \ref{eq: rope_theta}, $d$ is the dimensionality of the embedding space. 
\begin{align}
\label{eq: rope_theta}
\Theta = \left\{ \theta_i = 10000^{-2(i-1)/d}, i \in \left[1, 2, ..., d/2\right] \right\}
\end{align}



\section{The Proposed Method}
\label{sec_alg}

This section introduces our proposed LSTAN-GRPE model, designed for the road-network graph $G = (V_G, E_G)$. The model predicts the instantaneous vehicle velocity $V(\tilde{\tau})$ at a given time instant $\tilde{\tau}$, utilizing velocity sensor data from each road segment $\tilde{s}_{j} \in E_G$. 

Let $V(\tilde{s}_{j}, \tilde{\tau}_x:\tilde{\tau}_y)$ represent the traffic data for one road segment from $\tilde{\tau}_x$ to $\tilde{\tau}_y$, and let $\mathbb{V}_I(\tilde{\tau}_x:\tilde{\tau}_y)$ denote the set of all such data for all $\tilde{s}_{j} \in E_G$. Specifically, the model aims to learn a forecasting function $F_I(\cdot)$ that predicts $T$ steps ahead as follows:
\begin{equation}\label{eq: forecast_formalization}
\hat{\mathbb{V}}_I(\tilde{\tau}+1:\tilde{\tau}+T)=F_I(\mathbb{V}_I(\tilde{\tau}-T+1:\tilde{\tau});\mbox{ }M_{^eA}(G))
\end{equation}
where $M_{^eA}(G)$ is the edge-adjacency matrix of $G$.
Our model utilizes spatio-temporal attention to comprehensively consider data relevance. This includes spatial considerations along individual road segments and temporal aspects across time sequences.



\begin{figure}[!t]
\centering
\ifonecolumn
   \includegraphics[width=0.7 \textwidth]{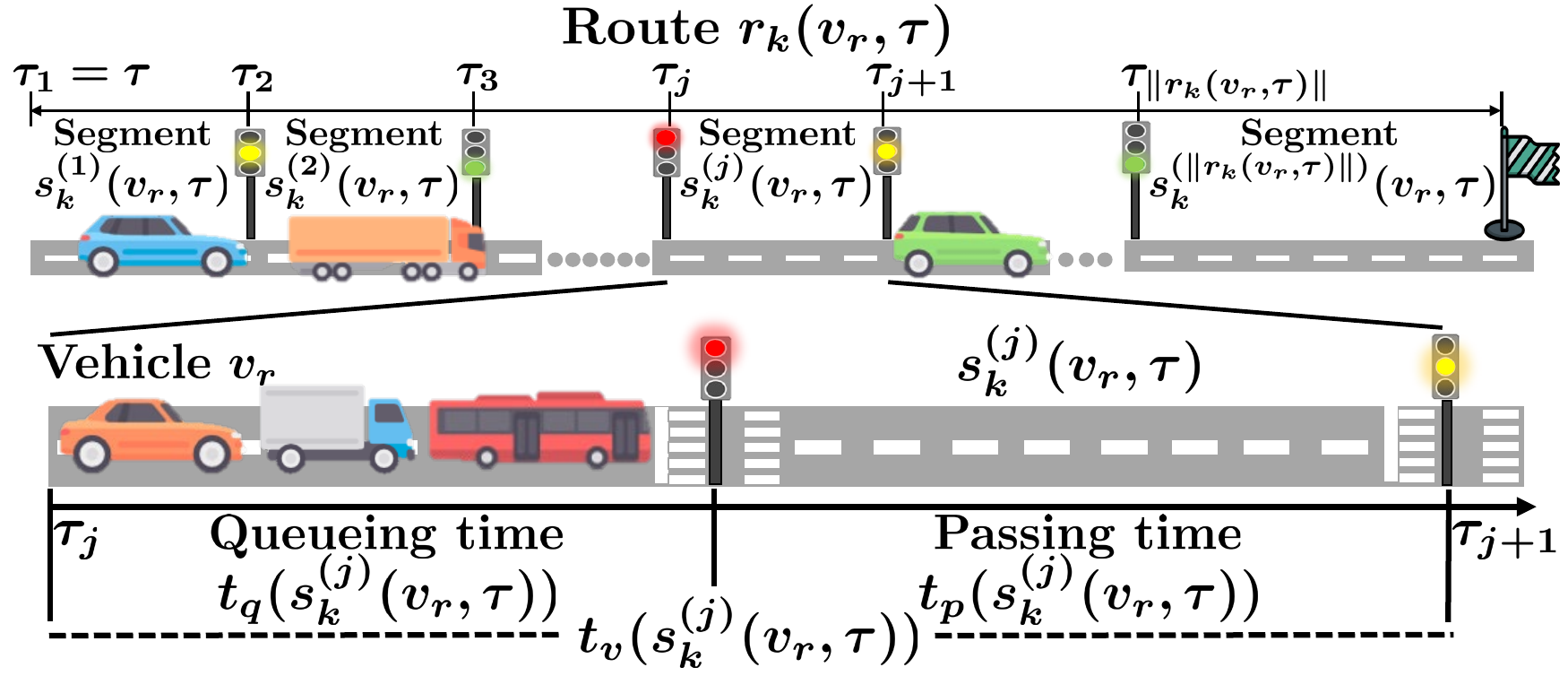}
    \else    
        \includegraphics[width=\columnwidth]{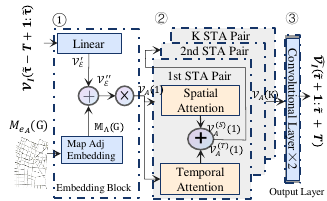} 
    \fi
\caption{Model Architecture}
    \label{fig:rercast_new}
\end{figure}



\subsection{Overall Architecture}
As depicted in Fig. \ref{fig:rercast_new}, our proposed model consists of a data embedding block \textcircled{1}, a sequential stack of spatio-temporal pairs \textcircled{2}, and the output layer \textcircled{3}. 

The initial data embedding block converts the input into a high-dimensional representation, yielding an intermediate value $V_{A}^{(1)}$, as shown in Fig. \ref{fig:rercast_new}.

Subsequently, we sequentially stack K pairs of spatio-temporal attention (STA) as shown in Fig. \ref{fig:rercast_new}. Recognizing that deeper networks can capture more complex features, thereby improving the model's accuracy and generalization capabilities, we stack $K$ spatio-temporal pairs in \textcircled{2}. In experiment, we set $K$ as $5$.
At $k$th STA pair, spatial attention and temporal attention operate in parallel.

Finally, the output layer utilizes two $1 \times 1$ convolutional layers which reshape and map the data from $T \times N \times \mathfrak{D}$ to $N \times T $, synthesizing the preceding spatio-temporal features for future T steps' traffic speed forecasting.

For the loss function, we adopt Huber Loss which blends the strengths of Mean Squared Error (MSE) and Mean Absolute Error (MAE) for optimal performance.

\subsection{Embedding Block}
At first, we apply a linear transformation to map the normalized traffic data $\mathbb{V}_I(\tilde{\tau}-T+1:\tilde{\tau}) \in \mathbb{R}^{N_s \times T}$ to $\mathcal{V}_I^{E'} \in \mathbb{R}^{N_s \times T \times \mathfrak{D}}$, where $\mathfrak{D}$ denotes the embedding dimension.

In the second step, we perform the eigenvalue decomposition on the normalized Laplacian matrix $M_L(G)$ of graph $G$ to embed graph features into $\mathcal{V}_I^{E'}$ to drive $\mathcal{V}_I^{E''}$. In particular, we use the eigenvectors of the graph Laplacian $M_L(G)$ to capture the structural information of the road network. Initially, we get the normalized Laplacian matrix $M_L(G)$ by:
\begin{equation}\label{eq: adj_laplacian}
\begin{aligned}
M_L(G) = M_I - {M_D(G)}^{-1/2}({M_{^eA}(G)}){M_D(G)}^{-1/2}
\end{aligned}
\end{equation}
where $M_I$ is the identity matrix, and $M_D(G)$ is the degree matrix of $G$. Subsequently, We use the decomposition of the eigenvalues of $M_L(G)$ by:
\begin{equation}\label{eq: eigenvector_laplacian}
\begin{aligned}
{M_U(G)}^{T} M_\Lambda(G) M_U(G) = M_L(G)
\end{aligned}
\end{equation}
to obtain the diagonal matrix $M_\Lambda(G)$ of eigenvalues and the eigenvector matrix $M_U(G)$.

Finally, we apply a linear projection to the eigenvector matrix $M_U(G)$ to derive the matrix $\mathbb{M}_U(G) \in \mathbb{R}^{N_s \times \mathfrak{D}}$, which is then added to $\mathcal{V}_I^{E'}$ to produce $\mathcal{V}_I^{E''}$:
\begin{equation}\label{eq: vb2}
\begin{aligned}
\mathcal{V}_I^{E''} = \mathcal{V}_I^{E'} + \mathbb{M}_U(G)
\end{aligned}
\end{equation}

The output of the embedding block $\mathcal{V}_{A}(1) \in T \times N_s \times \mathfrak{D}$.

\subsection{Spatio-Temporal Attention Pair}
We design the STA pair to simultaneously assess spatial and temporal dependencies in real time. For each \( k \in \{1, 2, \dots, K\} \), the $k$th STA pair comprises a spatial attention module and a temporal attention module, both processing the same input data $\mathcal{V}_{A}(k) \in \mathbb{R}^{N_s \times T \times \mathfrak{D}}$. $\mathcal{V}_{A}(k)$ will be fed separately to the temporal attention module and the spatial attention module. Subsequently, the STA pair assimilates the outputs from each module by additional fusion, and finally outputs $\mathcal{V}_{A}(k+1) \in \mathbb{R}^{N_s \times T \times \mathfrak{D}}$ as below:
\begin{equation}\label{eq: fusion}
\begin{aligned}
\mathcal{V}_{A}(k+1) = \mathcal{V}_{A}^{(S)}(k) + \mathcal{V}_{A}^{(T)}(k)
\end{aligned}
\end{equation}

\begin{figure}[!t]
\centering
\ifonecolumn
   \includegraphics[width=0.7 \textwidth]{pic/Traffic_Information_Prediction/traveltimesegments.pdf}
    \else    
        \includegraphics[width=\columnwidth]{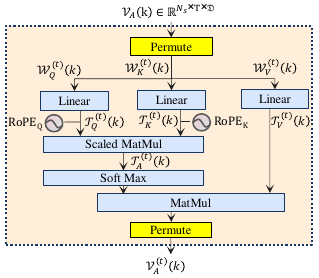} 
    \fi
\caption{Spatial / Temporal
Attention}
    \label{fig:stmodel}
\end{figure}

\subsection{Rotary Positional Encoding}
\label{sec: Rotary Positional Encoding}
To differentiate between spatial and temporal attention modules, we introduce the variable $t \in (S, T)$. Building upon the foundational concepts introduced in Section \ref{sec: Positional Encoding}, we first initiate the process by constructing a unified frequency sequence \(F^{(t)}\), where the variable $\Theta^{(t)}$ serves as a hyperparameter that controls the maximum frequency of the rotary encoding.

\begin{equation}
    F^{(t)} = \left\{ \pi \cdot \frac{2i - 1}{\mathfrak{D} - 1} \cdot \frac{\Theta^{(t)}}{2} \mid i = 1, 2, \ldots, \frac{\mathfrak{D}}{2} \right\}
\end{equation}
We then define two position sequence $P_T$ and $P_{N_S}$ as a linearly spaced sequence from $-1$ to $1$.
\begin{equation}
    P_T = \left\{ -1 + \frac{2i - 1}{T - 1} \mid i = 1, 2, \ldots, T \right\}
\end{equation}
\begin{equation}
    P_{N_S} = \left\{ -1 + \frac{2i - 1}{N_S - 1} \mid i = 1, 2, \ldots, N_S \right\}
\end{equation}

The mixed frequency $F^{(t)}_M$ is then computed using Equation \ref{eq:mixfrequency}. Specifically, we calculate the outer products of $P_T$ and $P_{N_S}$ with $F^{(t)}$ separately. These results are subsequently reshaped into three dimensions using the $Reshape^{(t)}(\cdot)$ function. For $t = S$, $Reshape^{(t)}(\cdot)$ transforms $P_T \otimes F^{(t)}$ into a tensor of shape $T \times 1 \times \frac{\mathfrak{D}}{2}$ and $P_{N_S} \otimes F^{(t)}$ into a tensor of shape $1 \times N_S \times \frac{\mathfrak{D}}{2}$. Conversely, for $t = T$, $Reshape^{(t)}(\cdot)$ transforms $P_T \otimes F^{(t)}$ into a tensor of shape $1 \times T \times \frac{\mathfrak{D}}{2}$ and $P_{N_S} \otimes F^{(t)}$ into a tensor of shape $N_S \times 1 \times \frac{\mathfrak{D}}{2}$. After broadcasting, the resulting shapes are $T \times N_S \times \frac{\mathfrak{D}}{2}$ for $t = S$ and $N_S \times T \times \frac{\mathfrak{D}}{2}$ for $t = T$. Finally, concatenation by function $Cat(\cdot)$ yields the mixed frequency tensors $F^{(S)}_M \in \mathbb{R}^{T \times N_s \times \mathfrak{D}}$ and $F^{(T)}_M \in \mathbb{R}^{N_s \times T \times \mathfrak{D}}$.
\begin{equation}
\label{eq:mixfrequency}
    \footnotesize F^{(t)}_M = Cat(Broadcast(Reshape^{(t)}(P_T \otimes F), Reshape^{(t)}(P_{N_S} \otimes F)))
\end{equation}

Finally, we define the rotary position embedding $ROPE^{(t)}(V)$ as
\begin{equation}
    \small ROPE^{(t)}(V) = V \cdot \cos(F^{(t)}_M) + RotateHalf(V) \cdot \sin(F^{(t)}_M)
\label{eq:ROPE}
\end{equation}
Here, the symbol $\cdot$ denotes the Hadamard product. The $RotateHalf(\cdot)$ function in RoPE involves splitting an input vector into two halves and then negating every other element in the second half, starting from the first element.

\subsection{Spatial and Temporal Attention}

The core motivation for spatial and temporal attention is to extract the similarity pattern from the spatial dimension and the temporal dimension. For spatial attention, since the city map is represented as a graph, with each segment considered as an edge within the graph's structure, the traffic data in road segments may exhibit similarity patterns. For temporal attention, as traffic data there exist sequential time series patterns, especially long range and periodic dependency relationship. Consequently, we design a spatial attention module to capture the data dependency between two distant road segments and a temporal attention module to capture the traffic data dependency along the time sequence.

The spatial attention module and temporal attention module share same module structure which is shown as Fig. \ref{fig:stmodel} except permutation calculation process which is only for spatial attention module. We reuse the variable $t$ in section \ref{sec: Rotary Positional Encoding}. As shown in the upper part of Fig. \ref{fig:stmodel}, the input to $k$th spatial Attention is $V_{A}^{(k)} \in \mathbb{R}^{N_s \times T \times \mathfrak{D}}$. if $t = S$, we permute $V_{A}^{(k)}$ to $\mathbb{R}^{ T \times N_s \times \mathfrak{D}}$, as we need to calculate the attention along the axis of road segment. if $t = T$, the permutation is omitted.

We follow the principle in \cite{vaswani2017attention} that Querys and keys are used to calculate similarity, which will then be applied to determine which Value should be focused on. We utilize the trainable parameters $\mathcal{W}_{Q}^{(t)}(k)$, $\mathcal{W}_{K}^{(t)}(k)$, $\mathcal{W}_{V}^{(t)}(k) \in \mathbb{R}^{\mathfrak{D} \times \mathfrak{D}}$ to multiply by $\mathcal{V}_{A}^{(t)}(k)$ separately, calculating $\mathcal{T}_Q^{(t)}(k), \mathcal{T}_K^{(t)}(k) , \mathcal{T}_V^{(t)}(k)$ which will be utilized in the attention model. The $\mathcal{T}_Q^{(t)}(k), \mathcal{T}_K^{(t)}(k) , \mathcal{T}_V^{(t)}(k) \in \mathbb{R}^{T \times N_s \times \mathfrak{D}}$ if $t = S$, and $\in \mathbb{R}^{N_s \times T \times \mathfrak{D}}$ if $t = T$.

\begin{equation}\label{eq: QKV_T}
\begin{aligned}
\mathcal{T}_Q^{(t)}(k) = RoPE^{(t)}(\mathcal{V}_{A}^{(t)}(k) \times \mathcal{W}_{Q}^{(t)}(k)) \\
\mathcal{T}_K^{(t)}(k) = RoPE^{(t)}(\mathcal{V}_{A}^{(t)}(k) \times \mathcal{W}_{K}^{(t)}(k)) \\
\mathcal{T}_V^{(t)}(k) = \mathcal{V}_{A}^{(t)}(k) \times \mathcal{W}_{V}^{(t)}(k)
\end{aligned}
\end{equation}
where function $RoPE^{(t)}(\cdot)$ is defined in Equation \ref{eq:ROPE}.

Subsequently, we compute the scaled spatial  self-attention score $\mathcal{T}_A^{(t)}(k)$ by two steps as follow.
\begin{equation}\label{eq: scale_matmul}
\begin{aligned}
  \mathcal{T}_A^{(t)}(k) =  \frac{\mathcal{T}_Q^{(t)}(k) \times (\mathcal{T}_K^{(t)}(k))^\top}{\sqrt{\mathfrak{D}}}   
\end{aligned}
\end{equation}
The first step is the production of the matrix of $\mathcal{T}_Q^{(t)}(k)$ and $\mathcal{T}_K^{(t)}(k)$. As $\mathcal{T}_Q^{(t)}(k)$ and $\mathcal{T}_K^{(t)}(k)$ are 3-D matrices, the multiplication to calculate vector similarity is performed across the last two dimensions along the spatial sequence.  Following this, we divide the production by $\sqrt{\mathfrak{D}}$ for scaling. This step benefits the stabilization of the training process and the control of the gradients magnitude. Then we yields $\mathcal{T}_A^{(t)}(k) \in \mathbb{R}^{ T \times N_s \times N_s}$, if $t = S$. And if $t = T$, $\mathcal{T}_A^{(t)}(k) \in \mathbb{R}^{ N_s \times T  \times T}$.

We then apply the softmax function to transform the scaled attention scores into weights.
These weights are subsequently used to perform a weighted sum of \(\mathcal{T}_V^{(t)}(k)\) to obtain the spatial self-attention output \(\mathcal{V}_{A}^{(t)}(k)\) as below:
\begin{equation}\label{eq: scale_matmul}
\begin{aligned}
\mathcal{V}_{A}^{(t)}(k) = Softmax(\mathcal{T}_A^{(t)}(k)) \times \mathcal{T}_V^{(t)}(k) 
\end{aligned}
\end{equation}
The $\mathcal{V}_{A}^{(t)}(k) \in \mathbb{R}^{T \times N_s \times \mathfrak{D}}$ if $t = S$, and $\mathcal{V}_{A}^{(t)}(k) \in \mathbb{R}^{N_s \times T \times \mathfrak{D}}$ if $t = T$.

Finally, by permuting \(\mathcal{V}_{A}^{(t)}(k) \in \mathbb{R}^{T \times N_s \times \mathfrak{D}}\), we obtain \(\mathcal{V}_{A}^{(t)}(k) \in \mathbb{R}^{N_s \times T \times \mathfrak{D}}\). If \(t = T\), permutation is skipped.

\section{Experiments}\label{sec:performance_evaluation}
\label{sec:Environment}
In this section, we present the experimental results of our proposed model for traffic flow forecasting. The analysis encompasses a comprehensive comparison of our model's performance with several recent benchmark models. Additionally, an ablation study is conducted to evaluate the contributions of key components within our network architecture, including spatial attention, temporal attention, RoPE, and graph eigenvector embedding.

\subsection{DataSet}
We evaluated the performance of ‌LSTAN-GERPE‌ using three real-world commonly used public traffic forecasting datasets: PeMS04 and PeMS08. Additionally, and their statistical information is summarized in TABLE \ref{tab:datasets}. 

\begin{table}[h]
\centering
\caption{Datasets Overview}
\label{tab:datasets}

\begin{tabular}{@{}>{\centering\arraybackslash}p{0.7cm}>{\centering\arraybackslash}p{0.6cm}>{\centering\arraybackslash}p{0.6cm}>{\centering\arraybackslash}p{1.1cm}>{\centering\arraybackslash}p{0.9cm}>{\centering\arraybackslash}p{2.7cm}@{}}
\toprule
Datasets & \#Nodes & \#Edges & \#Timesteps & \#Time Interval & Time range \\ \midrule
PeMS04 & 307 & 340 & 16992 & 5min & 01/01/2018-02/28/2018 \\
PeMS08 & 170 & 295 & 17856 & 5min & 07/01/2016-08/31/2016 \\ 
 \bottomrule
\end{tabular}
\end{table}

\subsection{Model and Data Settings}
In our research, we employ a standard approach to dataset partitioning, allocating 60\% of the data to the training set, 20\% to the validation set, and the remaining 20\% to the test set. Subsequently, we reshape the data by utilizing the velocity of the preceding 12 steps (equivalent to 60 minutes) as the input feature, with the objective of forecasting velocities for the next 12 steps.

All experiments are conducted on a system equipped with an NVIDIA GeForce RTX 4090 GPU, which features 24GB of dedicated GPU memory. The implementation of LSTAN-GRPE is carried out using Ubuntu 22.04, PyTorch version 2.0.1+cu118, and Python 3.9.16. The hidden dimension, denoted as $\mathfrak{D}$, is optimized by evaluating various values within the set $\left\{32,64,128,256,512\right\}$. Additionally, the depth of the encoder layers and the depth of the encoder layers $K$ is fine-tuned by considering values in $\left\{4,6,8,10,12,14\right\}$. The parameter $\Theta^{(t)}$ is also fine-tuned, with values considered from the set in $\left\{64, 128, 256, 512\right\}$. The model that exhibited the best performance is selected based on its performance in the validation data set.

For the training phase, we utilize the AdamW optimizer with a learning rate of 0.001. Training is conducted using a batch size of 16 over 200 epochs. An early stopping mechanism is implemented to terminate training if performance does not improve after reaching a predefined number of epochs without improvement.





\subsection{Ablation Study}
\begin{figure}[htbp]
    \centering
    \includegraphics[width=0.48\textwidth]{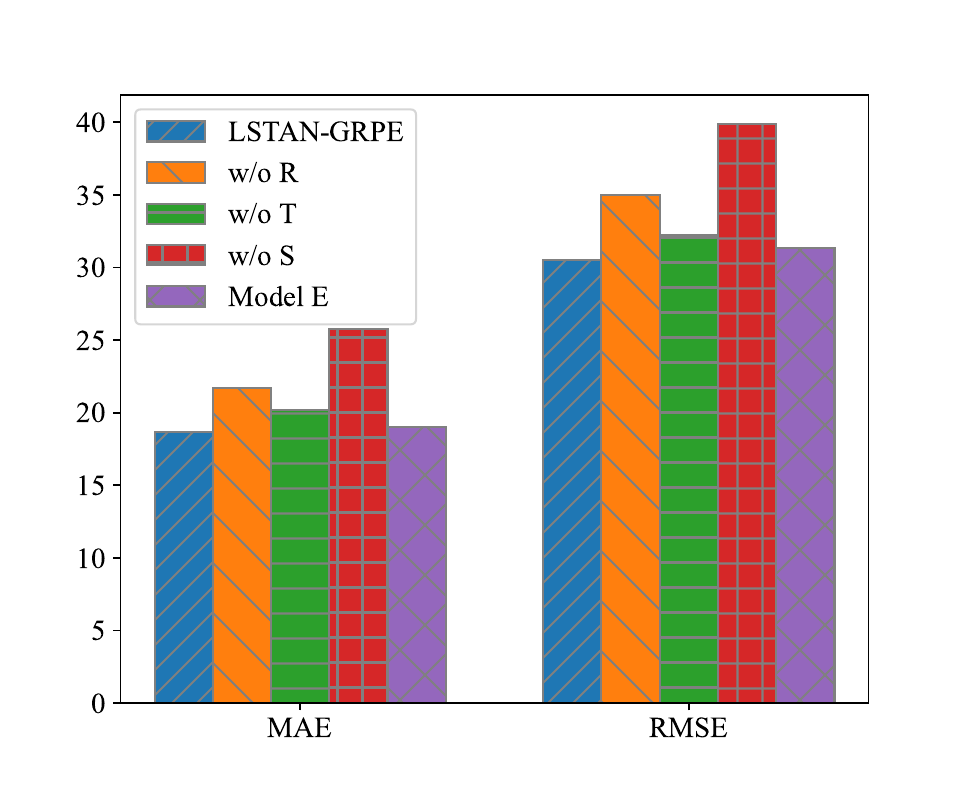}
    \caption{Ablation Study on Pems04}
    \label{fig:Ablation}
\end{figure}

An ablation study was conducted to investigate the significance of various components of our spatio-temporal model using the Pems04 dataset. By systematically removing or modifying components, we analyzed their contributions, as illustrated in Figure \ref{fig:Ablation}. The variants compared in this study included: (1) the complete model, (2) the model without rotary positional embedding (w/o R), (3) the model without temporal attention (w/o T), (4) the model without spatial attention (w/o S), and (5) the model without graph embedding (w/o E). 

The results of the ablation study are summarized as follows: (1) the complete model demonstrates the best performance, showing its overall effectiveness; (2) RoPE at an appropriate frequency is important, as it helps the model better understand the sequential data pattern; (3) spatial attention is found to be more significant than temporal attention, emphasizing the importance of spatial relationships in the data; (4) feature engineering by graph embedding can also help reduce the model's MSE and RMSE error metrics.

\subsection{Performance Comparison}
In our comparative analysis, LSTAN-GRPE was benchmarked against twelve baseline models, which can be categorized into three distinct classes.

Firstly, within the domain of traditional time series methods, the Vector Autoregression (VAR) model \cite{lu2016integrating} was included as a representative.

Secondly, among Graph Neural Network (GNN)-based models, we evaluated the Dynamic Convolutional Reaction Network (DCRNN) \cite{DBLP:conf/iclr/LiYS018}, Spatio-Temporal Graph Convolutional Network (STGCN) \cite{yu2018spatio}, Spatio-Temporal Synchronous Graph Convolutional Network (STSGCN) \cite{song2020spatial}, Spatio-Temporal Graph ODE Network (STGODE) \cite{fang2021spatial}, and Multi-scale Temporal Graph Neural Network (MTGNN) \cite{xie2024evaluations}.

Lastly, in the category of attention-based methods, we assessed the Spatial-Temporal Embedded Self-Attention Hybrid Network (STESHTN) \cite{tian2024hybrid}, Long-range Graph Feedforward Network (LLGformer) \cite{jin2025llgformer}, Spatio-Temporal Multi-Graph Convolutional Network (TMGCN) \cite{liu2024traffic}, Weight-Oriented Adaptive Graph Convolutional Recurrent Temporal Network (WOAAGCRTN) \cite{zhang2024adaptive}, Unified Spatio-Temporal Feature Attention Network (STFAN) \cite{guo2025unifying}, Enhanced Spatio-Temporal Graph Convolutional Network (STA-GCN) \cite{lu2024enhancing}, and Traffic Transformer \cite{yan2021learning}.

The comparative analysis presented in table \ref{tab:omparison} provides a detailed evaluation of various models on the PeMS04 and PeMS08 datasets, focusing on three key performance metrics: MAE, Mean Absolute Percentage Error (MAPE), and Root Mean Square Error (RMSE).  Our ‌LSTAN-GERPE demonstrates superior performance, highlighting its effectiveness in handling the complexities of traffic data prediction tasks.

\begin{table}[ht]
\centering
\scalefont{0.8} 
\caption{Performance comparison of different models on PeMS04 and PeMS08 datasets}
\label{tab:omparison}
\begin{tabular}{>{\centering\arraybackslash}p{2cm}>{\centering\arraybackslash}p{0.6cm}>{\centering\arraybackslash}p{0.8cm}>{\centering\arraybackslash}p{0.7cm}|>{\centering\arraybackslash}p{0.4cm}>{\centering\arraybackslash}p{0.8cm}>{\centering\arraybackslash}p{0.5cm}}
\toprule
\multirow{2}{*}{\centering\textbf{Model}}  & \multicolumn{3}{c|}{\textbf{PeMS04}} & \multicolumn{3}{c}{\textbf{PeMS08}} \\
\cmidrule(lr){2-4} \cmidrule(lr){5-7}
 & \textbf{MAE} & \textbf{MAPE(\%)} & \textbf{RMSE} & \textbf{MAE} & \textbf{MAPE(\%)} & \textbf{RMSE} \\
\midrule
VAR & 23.750 & 18.090 & 36.660 &  22.320 & 14.470 & 33.830 \\
\midrule
DCRNN & 21.92 & 14.46 & 35.01 & 17.5 & 11.08 & 27.29 \\
STGCN & 21.46 & 13.89 & 34.16  & 17.45 & 11.1 & 26.97 \\
STSGCN & 21.19 & 13.89 & 33.65 & 17.13 & 10.96 & 26.79 \\
STGODE & 20.84 & 13.78 & 32.82 &  16.81 & 10.62 & 26.1 \\
MTGNN & 18.87 & 13.45 & 30.97 & 15.31 & 10.21 & 24.25 \\

\midrule
TMGCN & 19.33 & 13.06 & 30.79 & 15.73 & 10.53 & 24.92\\
WOAAGCRTN & 19.13 & 12.77 & 31.37 & 15.27 & 9.96 & 24.67 \\
STA-GCNT & 19.37 & 12.69 & 30.58 & 15.24 & 9.96 & 24.83 \\
STESHTN & 19.61 & 12.91 & 31.64 & 15.92 & 10.96 & 25.6\\
LLGFormer & 19.12 & 12.88 & 30.59 & 15.16 & 10.08 & 24.21\\
Traffic-Transformer & 18.92 & 12.71 & 31.34 & 15.19 & 9.93 & 24.88 \\
STFAN & 19.79 & \textbf{12.7} & 31.86 & 15.56 & \textbf{9.8} & 24.57 \\
\midrule
‌LSTAN-GERPE‌ & \textbf{18.65} & 12.85 & \textbf{30.53} & \textbf{14.90} & 10.04 & \textbf{24.33} \\
\bottomrule
\end{tabular}
\end{table}

\section{Conclusion} \label{sec:conclusion}
In this study, we introduced LSTAN-GERPE, a novel lightweight spatio-temporal attention network for traffic forecasting. Our model leverages pure attention mechanisms to effectively capture both spatial and temporal dependencies. We integrated three key innovations: a lightweight spatio-temporal attention pair, RoPE with suitable frequency, and eigenvalue decomposition of the normalized Laplacian matrix for map graph.

Despite these advances, there are opportunities for further refinement. Future work could explore more sophisticated feature engineering techniques to enhance model performance. Additionally, applying our model to other spatio-temporal data types could broaden its utility.



\bibliographystyle{ieeetr}
\bibliography{references}

\end{document}